\documentclass{article}
\usepackage{ijcai22}

\usepackage{times}

\usepackage{soul}
\usepackage{url}
\usepackage[hidelinks]{hyperref}
\usepackage[utf8]{inputenc}
\usepackage[small]{caption}
\usepackage{graphicx}
\usepackage{amsmath}
\usepackage{booktabs}
\usepackage{multirow}
\usepackage{algorithm}
\usepackage{algorithmic}
\title{Local Hypergraph-based Nested Named Entity Recognition as Query-based Sequence Labeling}

\author{
Yukun Yan$^1$\And
Sen Song$^2$
\affiliations
$^1,^2$Laboratory of Brain and Intelligence, Tsinghua University, Beijing 100084, China\\
$^1,^2$Department of Biomedical Engineering, Tsinghua University, Beijing 100084, China\\
\emails
yanyk13@mails.tsinghua.edu.cn,
songsen@tsinghua.edu.cn
}

\begin{document}

\maketitle

\begin{abstract}
There has been a growing academic interest in the recognition of nested named entities in many domains. We tackle the task with a novel local hypergraph-based method: We first propose start token candidates and generate corresponding queries with their surrounding context, then use a query-based sequence labeling module to form a local hypergraph for each candidate. An end token estimator is used to correct the hypergraphs and get the final predictions. Compared to span-based approaches, our method is free of the high computation cost of span sampling and the risk of losing long entities. Sequential prediction makes it easier to leverage information in word order inside nested structures, and richer representations are built with a local hypergraph. Experiments show that our proposed method outperforms all the previous hypergraph-based and sequence labeling approaches with large margins on all four nested datasets. It achieves a new state-of-the-art F1-score on the ACE 2004 dataset and competitive F1-scores with previous state-of-the-art methods on three other nested NER datasets: ACE 2005, GENIA, and KBP 2017.\end{abstract}

\section{Introduction}

\begin{figure}[h]
	\centering
	\includegraphics[width=0.5\textwidth]{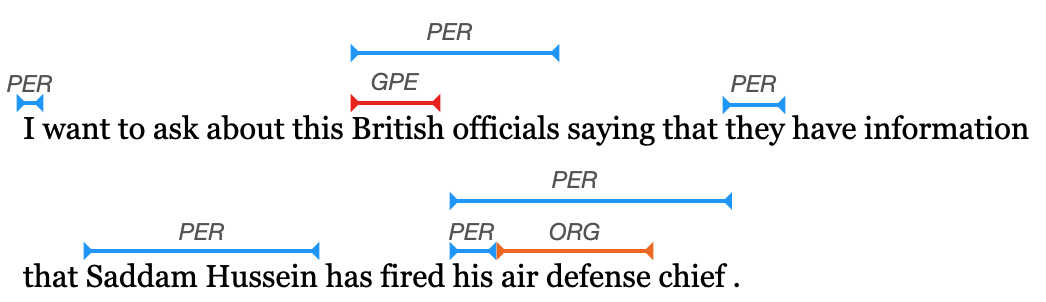}
	\caption{Examples of Nested Named Entity Recognition from ACE2005 dataset.}
	\label{fig:example}
\end{figure}

Named Entity Recognition (NER) is a fundamental task in natural language processing. It provides entity information for several downstream applications like coreference resolution and entity linking. Previous methods have achieved significant successes in flat named entity recognition by formulating it as a sequence labeling task. However, it is very common for entities to have nested structures, as shown in Figure \ref{fig:example}, which traditional approaches cannot handle since they only assign one tag to each token. 

To tackle this issue, multiple paradigms have been proposed in recent years. The span-based methods \cite{sohrab-miwa-2018-deep}\cite{tan2020boundary} first propose a certain number of span candidates that are allowed to be overlapped, and then predict their categories. However, span sampling brings a requirement of high computation cost, and comes with a risk of losing long entities. Besides, without the interactions between different candidates, the dependencies between entities are ignored. As for sequence labeling methods for Nested NER \cite{alex2007recognising}\cite{ju2018neural}\cite{wang2020pyramid}, previous works handle the nested structure by leveraging multiple decoding layers to identify outer entities and inner entities in a certain order. However, in the named entity recognition task, the output labels from these methods are essentially an unordered set. Another type of method \cite{lu-roth-2015-joint}\cite{katiyar-cardie-2018-nested} attempts to map a text into a single large hypergraph to capture all the named entities. Although they leverage structure information, these methods did not achieve competitive performance in recent years due to the overly complex designs. In this work, we focus on combining sequence labeling method with a simpler local hypergraph to recognize entities with nested structures. 

\begin{figure}[h]
	\centering
	\includegraphics[width=0.5\textwidth]{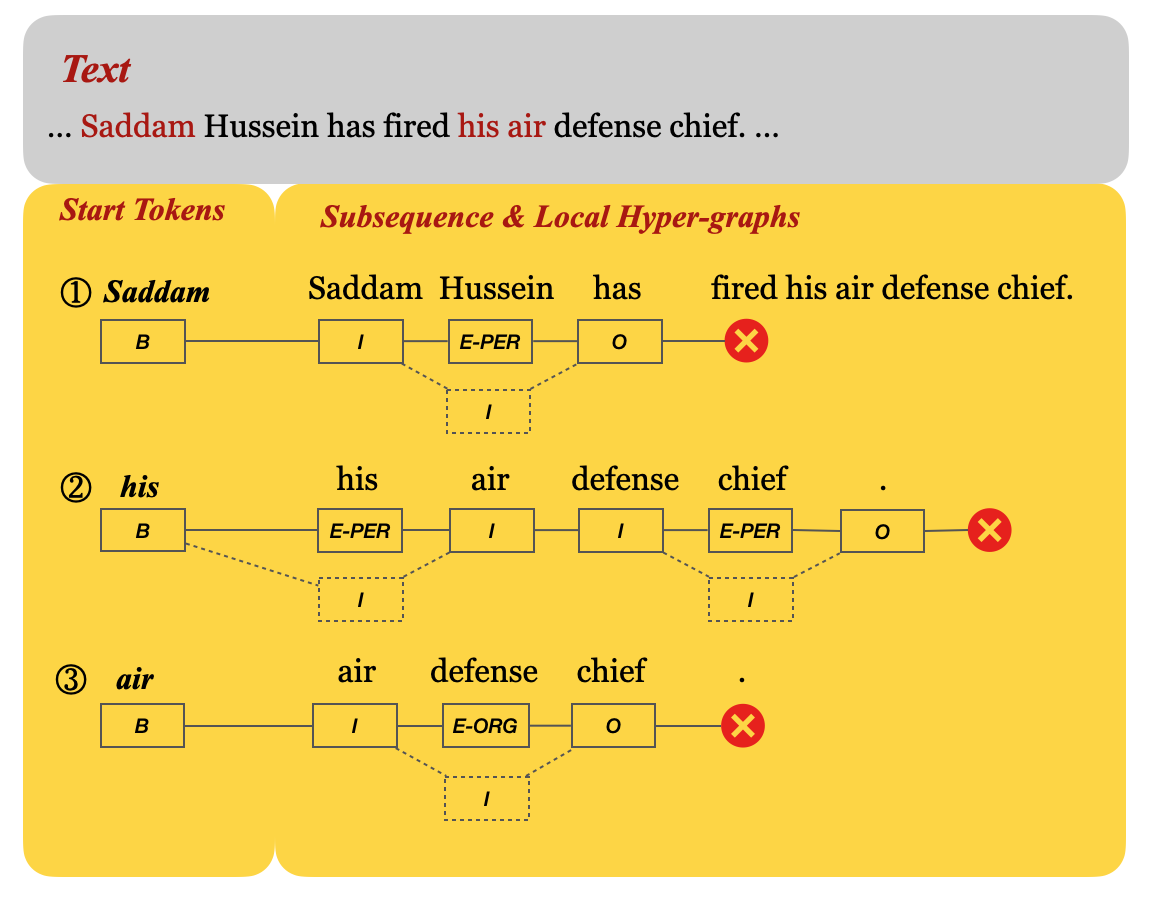}
	\caption{Proposed Local Hypergraph: Each row corresponds to a start token, its sub-sequence and hyper-graph. The entity information is represented with the well known B (beginning of entity), I (inside an entity), E-[cls] (end token of an entity of type [cls]), and O (outside any entity) tagging schema.}
	\label{fig:hypergraph}
\end{figure}

To handle nested structures, we introduce a local hypergraph as illustrated in Figure \ref{fig:hypergraph}. By the word local we emphasize that, unlike previous hypergraph-based methods, we capture the named entities by multiple hypergraphs corresponding to different sub-sequences instead of mapping the entire text into a complex full-size one. We observe that our brains recognize entities in the following process: we first identify the headwords which are usually at the boundary positions and then sequentially estimate whether an adjacent token can be merged. Inspired by this, we propose a novel method that treats the named entity recognition as building local hypergraphs by query-based sequence labeling. In general, we first sample a certain number of potential start tokens as the starting nodes of local hypergraphs. Given a start node, we sequentially tag the consecutive tokens as specific types of nodes to the current hypergraph. At last, the tokens recognized as end nodes are verified by an End Token Estimator to further correct the hypergraphs. Compared to the span-based methods, we set no length limit to entity candidates, and we reduce the time complexity from $O(n^2)$ to $O(n)$ by sampling tokens instead of spans.  Besides, sequential prediction makes it easier to leverage information in word order inside the nested structure which is ignored by span-based methods. The process of building the local hypergraph we use is much simpler than previous hypergraph-based approaches. In addition, with the structured information in local hypergraphs, our sequence labeling module is able to tag nested entities with only one decoding layer, and unlike previous sequence labeling methods, there is no need to decode entities in a specific order in our model. 

Our main contributions are as follows:

\begin{itemize}
	\item We introduce a novel local hypergraph, capable of representing entities with nested structures with no length limit. The construction process is much simpler than mapping a text into a single full-size hypergraph as used in previous studies.
	\item To our best knowledge, we are the first to formulate nested named entities as first proposing start positions and then constructing local hypergraphs by query-based sequence labeling. 
	\item Compared to previous sequence labeling approaches, our method is capable of tagging named entities of any length by only one decoding layer and does not have to follow a certain recognizing order.
	\item Compared to span-based methods, We samples start tokens instead of span candidates which reduces the time complexity of sampling process from $O(n^2)$ to $O(n)$. 
	\item Experiments show that our proposed method outperforms all of the previous hypergraph-based and sequence labeling methods. It achieves a new state-of-the-art F1-score on ACE 2004 dataset and competitive F1 scores with the best span-based models on the other three nested NER datasets: ACE2005, GENIA, and KBP2017.
\end{itemize} 

We will release our code upon publication of the paper.

\section{Related Work}
There are various paradigms for nested named entity recognition (NER). We can roughly divide them into span-based, hypergraph-based, sequence labeling methods, and other approaches. 

\textbf{Span-based Method} The span-based approaches are the most mainstream way for nested NER. Generally, they first propose a certain number of span candidates and then classify them into different categories. These studies focus on span sampling strategies and span representation methods. As an early attempt, Exhaustive Model \cite{sohrab-miwa-2018-deep} samples all possible spans. \cite{tan2020boundary} leverages sub-modules to estimate boundaries before sampling spans. \cite{shen2021locate} introduces a new module to adjust boundaries of span candidates to further use boundary information. \cite{DBLP:journals/corr/abs-2012-08478} uses a TreeCRF to enhance the interactions between nested spans. The most recent method \cite{yuan2021fusing} proposes tri-affine mechanism to integrate all useful information of different formats including tokens, labels, boundaries, and related spans to enhance the span representation. However, both the span proposing process and cross-span attentions bring a requirement of high computation cost. 

\textbf{Sequence Labeling Method} Previous sequence labeling methods for Nested NER\cite{alex2007recognising}\cite{luo2020bipartite}\cite{wang2020pyramid}\cite{shibuya2020nested} use multiple decoding layers to handle nested structures. The identification of entities in these methods follows a certain order, such as from inner to outer or from bottom-up, which is difficult to learn because the labels of entities are unordered.

\textbf{Hypergraph-based Method} Most previous hypergraph-based methods, like \cite{lu-roth-2015-joint}, are rule-based, and they attempt to map a text into carefully designed hypergraphs to capture all possible nested structures. Although \cite{katiyar-cardie-2018-nested} leverages a hypergraph to transform Nested NER to a modified sequence labeling task as we do in this work. This model tries to map a text to a single complex structure, and due to the complexity of the hypergraph, the model is required to do multiple binary classifications at each position, making it difficult to train. Hypergraph-based models did not achieve competitive performance in recent years. 

There are other methods for nested NER.  \cite{DBLP:journals/corr/abs-2105-08901} provides a fixed set of learnable vectors to learn the patterns of the valuable spans. \cite{li-etal-2020-unified} uses a machine reading comprehension framework to identify entities. However, additional data is required in this work to generate query representation for each entity type.  

To our best knowledge, we are the first to formulate nested named entities recognition as a query-based sequence labeling task by first proposing start positions and then constructing local hypergraphs from them. 

\begin{figure*}[h!]
	\centering
	\includegraphics[width=0.75\textwidth]{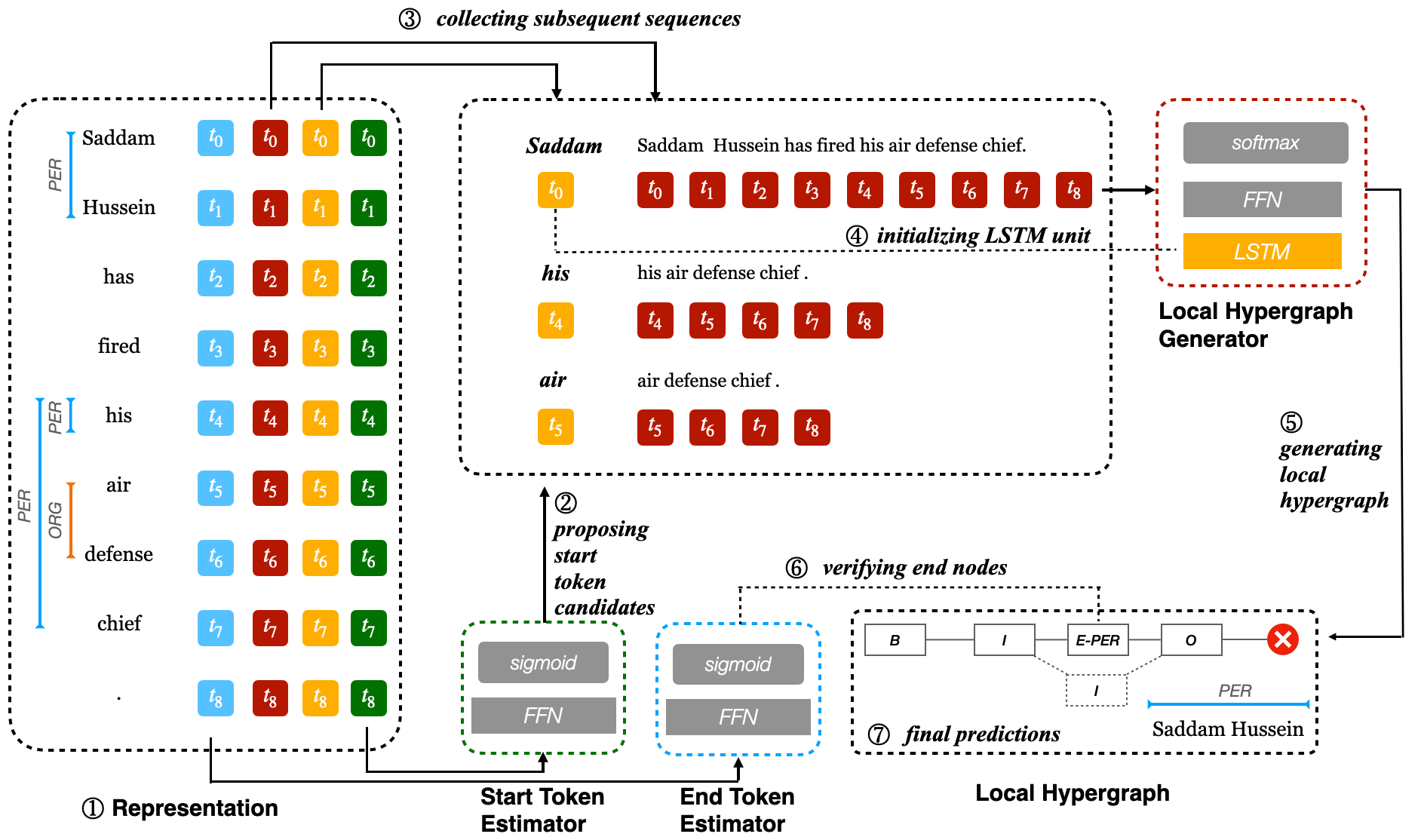}
	\caption{The paradigm of our method: \textcircled{1} Generating token representations with four encoders; \textcircled{2} Proposing potential start tokens; \textcircled{3} collecting corresponding sub-sequences; \textcircled{4} Initializing the cell memory and hidden state of the LSTM unit in the Local Hypergraph Generator with proposed start tokens; \textcircled{5} Generating local hypergraph; \textcircled{6} Verifying end nodes by End Token Estimator; \textcircled{7} Final predictions from local hypergraphs}
	\label{fig:method}
\end{figure*}

\section{Method}

Figure \ref{fig:method} illustrates an overview of our methods. There are three main modules in our approach, referred to as \textbf{Start Token Estimator}, \textbf{Local Hypergraph Generator}, and \textbf{End Token Estimator}, each corresponding to a processing stage. 

At the first stage, we use the Start Token Estimator to assign a score to each token to evaluate the likelihood that it is a start token. We select a certain number of tokens with the highest scores and regard them as candidates for the next stage. At stage two, we collect all the start token candidates and their subsequent sequences. We encode a start token candidate $t^*$ together with its context to form a representation of the query \textsl{`With current context, what is the hypergraph that starts with the $t^*$?'}, which is then used to initialize the Local Hypergraph Generator. It generates a tag sequence for the corresponding subsequent sequence to form a local hypergraph until an O-node is predicted or all the tokens are tagged. At the final stage, the End Token Estimator produces a score for each token tagged as an E-node to estimate whether it is an end token. By setting a threshold, we further revise the local hypergraph by removing the obvious faulty E-node. As the process is terminated by the O-node, this does not affect the overall length of a local hypergraph, but only the number of entities identified.

Specifically, in our approach, there are four individual encoders with the same structure that separately provide token representations with one for the Start Token Estimator, one for the End Token Estimator, and two for the Local Hypergraph Generator. The encoder is based on BERT and BiLSTM. The Start and End Token Estimators have the same structure that consists of a two-layer feedforward network. The Local Hypergraph Generator has a one-layer LSTM and a two-layer feedforward network.

The detailed information about the modules is described below.
%\begin{algorithm}
%\caption{Local Hypergraph Generation with Tag Sequence}
%$initialization$\\
%$N = [B]$\\
%\While{$i \neq n$ and $T_i \neq O$}
%{
% float $maxSim=0$\;
% $r^{maxSim}=null$\;
%  \While{not end of $T_j$}
% {
%     compute Jaro($r_i,r_m$)($r_m \in T_j$)\;
%     \If{$(Jaro(r_i,r_m) \ge \theta _r) \wedge ((Jaro(r_i,r_m) \ge r^{maxSim}) $}
%      {
%           replace $r^maxSim$ with $r_m$\;
%      }   
% }
%$con(r_i)=con(r_i) \cup {r^{maxSim}}$\;
%}
%return $con(r_i)$\;
%\end{algorithm}

%We describe the construction process of the second example in Figure \ref{fig:hypergraph} below to better illustrate the local hypergraph:
%\begin{itemize}
%	\item With the token \'his\' as a start token candidate, the hypergraph is initialized with a single B-node.
%	\item Since the first token \'his\' in the subsequent sequence is tagged as an E-PER-node, we add an E-PER-node and an I-node to the B-node.
%	\item The second token \'air\' is 
%\end{itemize} 

\subsection{Text Encoding}

Given a text with $n$ tokens, the representation of the $i$-th token is built by  first concatenating four components: (1) the encoded result $t_i^{lm}$ of a language model; (2) word embedding $t_i^w$; (3) part-of-speech (POS) embedding $t_i^{pos}$; and (4) character-level embedding $t_i^{char}$ which is generated by a BiLSTM model. Then, we fed them into a BiLSTM layer and regard its hidden states as the token representation $t_i$. The encoding method is the same as \cite{shen2021locate}.
\begin{equation}
	z_i = t_i^{lm} \oplus t_i^w \oplus t_i^{pos} \oplus t_i^{char}
\end{equation}
\begin{equation}
\overrightarrow{h_i} = \overrightarrow{LSTM}([z_0, z_1, ..., z_n])
\end{equation}
\begin{equation}
\overleftarrow{h_i} = \overleftarrow{LSTM}([z_0, z_1, ..., z_n])
\end{equation}
\begin{equation}
t_i = \overrightarrow{h_i} \oplus \overleftarrow{h_i}
\end{equation}

Considering that there are four individual encoders, to better describe following sections, we label the generated results with different superscripts to distinguish different usages: $T^s = [t_1^s, t_2^s ...]$ is the input of the Start Token Estimator; $T^e = [t_1^e, t_2^e ...]$ is the input of the End Token Estimator; $T^q = [t_1^q, t_2^q ...]$ is used as the representation of queries used by the Local Hypergraph Generator;  $T^c = [t_1^c, t_2^c ...]$ is used to represent sequence content for the Local Hypergraph Generator.

\subsection{Tokens Estimators}

As a binary classifier, the Start Token Estimator is trained to propose start token candidates at processing stage one. Similar to the proposing strategy of span-based methods, we prefer to obtain a high-recall sampling result. Thus, we use a focal loss with $\gamma = 0.9$ as our target function to minimize, and the same configuration goes for the End Token Estimator.

\begin{equation}
p_i^{s/e} = sigmoid(FFN(t_i^{s/e}))
\end{equation}
\begin{equation}
L_{cls}^{s/e} = \sum_{i=1}^n{- y_i^{s/e} \cdot (1 - p_i^{s/e})^\gamma log(1 - p_i^{s/e})}
\end{equation}

where $p_i^{s/e}$ is the score to evaluate the likelihood of $t$-th token being a start/end token; $L_{cls}^{s/e}$ is the loss function, and $y_i^{s/e}$ is the true label of the $i$-th token.

The sampling strategy of start tokens during training is important. If we only use high-quality candidates during training, the model will have trouble dealing with low-quality start tokens during evaluation. On the other hand, too many false positive tokens tend to disturb the learning process. Thus, we introduce a multiplicative constant $\lambda$ in start token candidates sampling. Specifically, if there are $n$ tokens that have a score higher than $0.5$, we collect $\lambda * n$ tokens with the highest scores as candidates. 

\subsection{Local Hypergraph}

To represent all possible nested structures, the local hypergraph we use has the following rules:

\begin{itemize}
	\item \textbf{Rule 1:} There are four types of nodes: (1) A B-node indicates a beginning token of an entity; (2) An I-node stands for a token inside an entity; (3) An E-[cls]-node represents the end token of an entity whose category is [cls]. (4) an O-node refers to a token outside an entity.
	\item \textbf{Rule 2:} (1) There is only one B-node in a hypergraph; (2) There is at most one O-node as the last node in a hypergraph; (3) There can be any numbers of I-nodes and E-[cls]-nodes.
	\item \textbf{Rule 3:} (1) If an E-[cls]-node is generated, an I-node is also added to the hypergraph. They share the same superior node. (2) If an O-node is connected to the hypergraph, it stops growing. (3) If a node is added to the hypergraph, it connects to all the nodes generated at the previous time step.
	\item \textbf{Rule 4:} If a path starts with a B-node, ends with an E-[cls]-node, and the other nodes (if any) are all I-nodes, it is labeled as a named entity of category [cls]. 
\end{itemize}

While the Local Hypergraph Generator sequentially generates tags one by one, a local hypergraph is generated in the following process: At the very beginning, a local hypergraph is initialized with a single B-node. At each time step, if an O-node is predicted, the hypergraph stops growing; If an I-node is predicted, we connect an I-node to the very last I-node or B-node; If an E-[cls]-node is predicted, an I-node and an E-[cls]-node is connected to the very last I-node or B-node. After the generation process is finished, all entities are labeled by the paths described in \textbf{Rule 4}.  

\subsection{Local Hypergraph Generator}

In stage two, in each sequence labeling process, we identify all the entities that start with the same token. We generate a local hypergraph for each start token by the Local Hypergraph Generator with a fragment starting with it as the input sequence. The generator consists of a one-layer LSTM and a feedforward layer. Since a token could be tagged with different node types in different local hypergraphs, as shown in the second and the third examples in Figure \ref{fig:hypergraph}, the information of the start token should be taken into consideration. To address the problem, given a start word $t^*$, we introduce a query-based sequence labeling method that initializes its state with a representation of the query \textsl{`With current context, what is the hypergraph that starts with the $t^*$?'}. Specifically, we initialize the cell memory and hidden state of the LSTM cell with the representation of the start token encoded with a separate encoder.  

\begin{equation}
C_0 = h_0 = t_{i^*}^q
\end{equation}
\begin{equation}
C_t =f_t \cdot C_{t-1} + i_t \cdot tanh(W_c \cdot [h_{t-1}] + b_c) 
\end{equation}
\begin{equation}
h_t = o_t \cdot tanh(C_t) 
\end{equation}
\begin{equation}
h_t = LSTM(h_{t-1}, t_t^c), \quad t_t \in [t_{i^*}^c, t_{i^* + 1}^c, ...]
\end{equation}
\begin{equation}
p_t = softmax(FFN(h_t))
\end{equation}
Where  $i_t$, $f_t$, $o_t$, $C_t$, and $h_t$ separately indicate the input gate, forget gate, output gate, cell memory, and hidden state of the LSTM layer. $p_t$ is the predicted probability of node types. 

We use cross-entropy cost as the target function $L^{hg}$ to minimize while training the Local Hypergraph Generator. Given $N$ start token candidates and its subsequent sequence consisting of $T_i$ tokens, the loss is computed as in equation (12).

\begin{equation}
L^{hg} = \frac{1}{N} \sum_{i=1}^N \sum_{t=1}^{T_i} CELoss(y_{t}, p_{t})
\end{equation}

Since the generation stops once an O-node is predicted, we limit the length of a subsequent sequence to only contain one O-node to save computation resources. 

During the evaluation, the subsequent sequence retrieved with each start token ends with the last token of a sentence. 

Since the generation process is the core of our algorithm, we provide pseudocode below for better description.
\begin{table*}[h] \footnotesize
	\begin{center}
	\resizebox{0.95\linewidth}{!}{
		\begin{tabular}{cccccccccccc}
			\hline 
			\multirow{2}*{Dataset Statistics} & \multicolumn{3}{c}{ACE 2004} & \multicolumn{3}{c}{ACE 2005} & \multicolumn{3}{c}{KBP 2017} & \multicolumn{2}{c}{GENIA}\\ 
			\cline{2-12}
			 ~& Train & Dev & Test & Train & Dev & Test & Train & Dev & Test & Train & Test\\
			\hline
			\# sentences & 6200 & 745 & 812 & 7194 & 969 & 1047 & 10546 & 545 & 4267 & 16692 & 1854\\
			\# with nested entities & 2712 & 294 & 388 & 2691 & 338 & 320 & 2809 & 182 & 1223 & 3522 & 446\\
			\# avg sentence length & 22.50 & 23.02 & 23.05 & 19.21 & 18.93 & 17.2 & 19.62 & 20.61 & 19.26 & 25.35 & 25.99\\
			\# total entities & 22204 & 2514 & 3035 & 24441 & 3200 & 2993 & 31236 & 1879 & 12601 & 50509 & 5506\\
			\# nested entities & 10149 & 1092 & 1417 & 9389 & 1112 & 1118 & 8773 & 605 & 3707 & 9064 & 1199\\
			nested percentage(\%) & 45.71 & 46.69 & 45.61 & 38.41 & 34.75 & 37.35 & 28.09 & 32.20 & 29.42 & 17.95 & 21.78\\
			\hline
		\end{tabular}}
	\end{center}
	\caption{Statistics of the datasets used in the experiments.}
	\label{t:data}
\end{table*}
\begin{algorithm}[tb]
\caption{The generation process of a local hypergraph}
\label{alg:algorithm}
\textbf{Input}: start token candidate $t_i$, and sequence $[t_i, t_{i+1} ..,t_{i+n}]$\\
\begin{algorithmic}[1] %[1] enables line numbers
\STATE Let $C_0 = h_0 = t^q_i$, $j=1$, $l_0=B-node$, $L = [l_0]$
\WHILE{$l_j \neq O-node$ and $j \leq n + 1$}
\STATE $h_j = LSTM(h_{j-1}, t_{i + j - 1}^c)$
\STATE $l_j=argmax(softmax(FFN(h_j)))$
\STATE $L.append(l_j)$
\STATE $j = j + 1$
\ENDWHILE
\STATE parsing predicted tag with \textbf{Rule 3} in section 3.3
\STATE \textbf{return} local hypergraph $G$
\end{algorithmic}
\end{algorithm}

\section{Experiments}
\subsection{Datasets}

To evaluate the proposed method, we conduct experiments on four widely used datasets for Nested NER: ACE2004, ACE2005, KBP2017 an GENIA. Detail statistics of the datasets used in the experiments are shown in Table \ref{t:data}.

\textbf{ACE 2004 and ACE 2005}\cite{doddington2004automatic}\cite{ace05} are nested datasets with 7 entity categories, we use the same setup as previous works\cite{katiyar-cardie-2018-nested}\cite{shen2021locate} and split them into train, dev, and test sets by 8:1:1.

\textbf{KBP 2017}\cite{ji2017overview} has 5 entity categories. We split all the samples into 866/20/167 documents for train/dev/test set following the same setup as previous works\cite{shen2021locate}.

\textbf{GENIA}\cite{ohta2002genia} is a nested dataset consisting of biology texts. There are 5 entity types: DNA, RNA, protein, cell line and cell categories. Following \cite{shen2021locate}, we use a 90\%/10\% train/test split.

\subsection{Evaluation Metrics}

We employ precision, recall and F1-score to evaluate the performance. Here we use strict evaluation metrics that an entity is considered correctly labeled only if its boundary and category are correct simultaneously.  

\subsection{Parameter Setting}
 
 In the experiments on \textbf{ACE 2004}, \textbf{ACE 2005} and \textbf{KBP 2017}, we leverage BERT\cite{devlin2018bert} and GloVE\cite{pennington2014glove} to initialize our encoders. The dimensions for $t_i^{lm}$, $t_i^w$, $t_i^{pos}$, $t_i^{char}$, and $t_i$ are 1024, 300, 256, 256, and 1836, respectively. We replace BERT and GloVe with SciBERT\cite{DBLP:journals/corr/abs-1903-10676} and BioWordvec\cite{chiu-etal-2016-train} for \textbf{GENIA}. Corresponding dimensions of $t_i^{lm}$, $t_i^w$ are 768, 200. Considering the large number of low-frequency words in the GENIA, we change the  dimension of $t_i^{char}$ to 1024. For all the experiments, we train our model for 80 epochs with an AdamW optimizer and a linear warmup-decay learning rate. The basic learning rate for BERT modules, pre-trained word embedding vectors, and other parameters are set to 1e-5, 1e-6, and 2e-4 respectively. The sampling parameter $\lambda$ is set to 3 for training and 1.5 for evaluation. The threshold of End Token Estimator is set to 0.2.
 
\subsection{Baselines}
We compare our method with several state-of-the-art approaches, including span-based, hypergraph-based, sequence labeling, and other methods, on ACE 2004, ACE 2005, KBP 2017 and GENIA datasets: 
\begin{itemize}
	\item \cite{DBLP:journals/corr/abs-2012-08478} uses a TreeCRF to enhance interactions between nested spans then predicts their types.
	\item \cite{tan2020boundary} proposes a boundary enhanced neural span classification model.
	\item \cite{shen2021locate} proposes a two-stage entity identifier to maintain high-quality span candidates.
	\item \cite{yuan2021fusing} proposes a novel tri-affine mechanism including tri-affine attention and scoring.
	\item \cite{lu-roth-2015-joint} proposes a hypergraph to jointly do mention extraction and classification.
	\item \cite{katiyar-cardie-2018-nested} makes use of the BILOU tagging scheme to learn the hypergraph representation.
	\item \cite{luo2020bipartite} proposes a bipartite flat-graph network with two interacting subgraph modules.
	\item \cite{shibuya2020nested} searches a span of each extracted entity for nested entities with second-best sequence decoding.
	\item \cite{wang2020pyramid} designs the normal and inverse pyramidal structures to identify entities through bidirectional interactions.
	\item \cite{DBLP:journals/corr/abs-2105-08901} provides a fixed set of learnable vectors to learn the patterns of the valuable spans.
\end{itemize}

\subsection{Overall Result}

\begin{table*}[ht] \footnotesize
	\begin{center}
	\resizebox{1\linewidth}{!}{
		\begin{tabular}{ccccccccccccc}
			\hline 
			\multirow{2}*{Models} & \multicolumn{3}{c}{ACE 2004} & \multicolumn{3}{c}{ACE 2005} & \multicolumn{3}{c}{KBP 2017} & \multicolumn{2}{c}{GENIA}\\ 
			\cline{2-13}
			 ~& Precision & Recall & F1-score & Precision & Recall & F1-score & Precision & Recall & F1-score & Precision & Recall & F1-score \\
			\hline
			\textbf{Span-based Methods}\\
			\cite{DBLP:journals/corr/abs-2012-08478} & 84.40 & 85.40 & 84.90 & 82.0 & 86.40 & 84.10 & - & - & - & 80.50 & 74.50 & 77.40\\ 
			\cite{tan2020boundary} & 85.80 & 84.80 & 85.30 & 83.80 & 83.90 & 83.90 & - & - & - & 79.20 & 77.40 & 78.30 \\
			\cite{shen2021locate} & 87.44 & 87.38 & 87.41 & 86.09 & 87.27 & 86.67 & 85.46 & 82.67 & 84.05 & 80.19 & 80.89 & 80.54\\
			\cite{yuan2021fusing} & 87.13 & 87.68 & 87.40 & 86.70 & 86.94 & 86.82 & 86.50 & 83.65 & 85.50 & 80.42 & 82.06 & 81.23 \\
			\hline
			\textbf{Hypergraph-based Methods}\\
			\cite{lu-roth-2015-joint} & 74.40 & 50.00 & 59.80 & 63.40 & 53.80 & 58.30 & - & - & - & 72.50 & 65.20 & 68.70\\
			\cite{katiyar-cardie-2018-nested} & 73.60 & 71.80.8 & 72.70 & 70.60 & 70.40 & 70.50 & - & - & - & 79.80 & 68.20 & 73.60 \\
			\hline
			\textbf{Sequence Labeling Methods}\\
			\cite{luo2020bipartite} & - & - & - & 75.00 & 75.20 & 75.10 & 77.10 & 74.30 & 75.60 & 77.40 & 74.60 & 76.00 \\
			\cite{shibuya2020nested} & 83.73 & 81.91 & 82.81 & 82.92 & 82.42 & 82.70 & - & - & - & 78.07 & 76.45 & 77.25 \\
			\cite{wang2020pyramid} & 86.08 & 86.48 & 86.28 & 83.95 & 85.39 & 84.66 & - & - & - & 80.33 & 78.31 & 79.31 \\
			\hline
			\textbf{Other Methods}\\
			\cite{DBLP:journals/corr/abs-2105-08901} & 88.46 & 86.10 & 87.26 & 87.48 & 86.63 & 87.05 & 84.91 & 83.04 & 83.96 & 82.31 & 78.66 & 80.44\\
			\hline
			\textbf{our model} & 88.15 & 88.30 & \textbf{88.23} & 87.61 & 87.33 & \textbf{87.40} & 86.8 & 85.27 & \textbf{86.03} & 83.34 & 80.78 & \textbf{82.03}\\
			\hline
		\end{tabular}}
	\end{center}
	\caption{Results on the nested datasets: ACE 2004, ACE 2005, KBP 2017, and GENIA.}
	\label{t:overall_result}
\end{table*}

The performance of the proposed method and baselines is shown in Table \ref{t:overall_result} on all four datasets. Our method outperforms all the state-of-the-art models on ACE 2004, achieves the second-best F1-score on KBP 2017, and acquires competitive results on the other two datasets. Especially, as the same type of methods, our method outperforms previous best sequence labeling and hypergraph-based methods with large margins on all four datasets. 

\subsection{Ablation Study}

We conduct ablation experiments on ACE 2004 to further elucidate the value of querying with the start token when building the hypergraph. Specifically, we use the following three settings and the result is illustrated in Table \ref{t:ablation_query}.

(a) We initialize the cell memory and hidden state of the LSTM unit with all zero vectors instead of the start token. 

(b) We use the same encoder for encoding the subsequent sequence into $T^c$ and generating the representation $T^q$ of the start token. 

(c) Our full model.

\begin{table}[ht] \footnotesize
	\begin{center}
	\resizebox{0.3\textwidth}{!}{
		\begin{tabular}{cccc}
			\hline 
			Setting & Precision & Recall & F1-score\\
			\hline
			(a) & 86.06 & 86.43 & 86.50\\
			(b) & 87.98 & 86.17 & 87.07\\
			(c) & 87.67 & 87.39 & \textbf{87.47}\\
			\hline
		\end{tabular}}
	\end{center}
	\caption{Ablation study on representation of queries.}
	\label{t:ablation_query}
\end{table}

From the result shown in Table \ref{t:ablation_query} we can conclude that the query plays an important role in the construction of local hypergraphs. Furthermore, comparing (b) with (c), we find that using a separate encoder to build a query is better.  

\begin{table}[ht] \footnotesize
	\begin{center}
	\resizebox{0.5\textwidth}{!}{
		\begin{tabular}{ccccccc}
			\hline 
			\multirow{2}*{threshold} & \multicolumn{3}{c}{ACE 2004} & \multicolumn{3}{c}{GENIA} \\
			~ & Precision & Recall & F1-score & Precision & Recall & F1-score\\
			\hline
			None & 87.44 & 87.41 & 87.42 & 81.97 & 77.67 & 79.76 \\
			0.1 & 87.57 & 87.40 & \textbf{87.48} & 82.34 & 77.49 & 79.84\\
			0.2* & 87.67 & 87.39 & 87.47 & 82.53 & 77.45 & \textbf{79.91}\\
			0.5 & 87.70 & 87.39 & 87.45 & 83.16 & 76.41 & 79.64\\
			0.8 & 88.20 & 86.21 & 87.19 & 84.33 & 73.21 & 78.38\\
			\hline
		\end{tabular}}
	\end{center}
	\caption{Ablation study on the confirming threshold of End Token Estimator. }
	\label{t:ablation_end_estimator}
\end{table}

To evaluate the value of the end estimator, we set different thresholds, and conduct experiments on ACE 2004 and GENIA datasets. The results is illustrated in Table \ref{t:ablation_end_estimator}, from which we can see that the End Token Estimator can help increase precision by removing faulty E-node. Comparing the results on ACE 2004 and GENIA, we found that it plays a more important role in the case of GENIA dataset. We took a closer look at the samples in these two datasets, and found that compared to ACE 2004, most headwords are at the end of the entities in GENIA, but the local hypergraphs are built from the start token. That partially explains the different performance gains of a separate evaluation for the end tokens.
\section{Conclusion}
By introducing a local hypergraph to handle nested structures, we propose a novel method that treats Nested NER as a query-based sequence labeling task. First, we propose a certain number of start token candidates, and then we generate a local hypergraph for each candidate with a query-based sequence labeling method. We get the final prediction after using an End Token Estimator to correct faulty end tokens. Our method has a significantly lower sampling complexity compared to span-based methods. Although we use a hypergraph, generating multiple local structures instead of mapping a whole text avoids an overly complex construction process. In addition, leveraging local hypergraph makes our sequence labeling module free from identifying entities in a certain order that the previous sequence labeling methods suffer from. Our method outperforms all the previous hypergraph-based and sequence labeling approaches with large margins on all four nested datasets. It achieves a new state-of-the-art F1-score on the ACE 2004 dataset and competitive F1-score with previous state-of-the-art span-based approaches on three other nested NER datasets: ACE2005, GENIA, and KBP2017. For future work, we will improve our method by bidirectionally constructing local hypergraphs, and try to build a more effective query representation.
 
\bibliographystyle{named}
\bibliography{ijcai22.bib}

\end{document}